\documentclass[runningheads]{llncs}
\usepackage[T1]{fontenc}
\usepackage{graphicx}
\usepackage[table,xcdraw]{xcolor}
\usepackage{booktabs}
\usepackage{tablefootnote}
\usepackage{multicol}
\usepackage{multirow}
\usepackage{caption}
\begin{document}
\title{Local LLM Ensembles for Zero-shot Portuguese Named Entity Recognition}
%
%
\author{João Lucas Luz Lima Sarcinelli \and
Diego Furtado Silva}
\authorrunning{J. L. L. L. Sarcinelli et al.}
\institute{Instituto de Ciências Matemáticas e Computação, Universidade de São Paulo
\email{\{joao.luz,diegofsilva\}@usp.br}}
\maketitle              
\begin{abstract}
Large Language Models (LLMs) excel in many Natural Language Processing (NLP) tasks through in-context learning but often under-perform in Named Entity Recognition (NER), especially for lower-resource languages like Portuguese. While open-weight LLMs enable local deployment, no single model dominates all tasks, motivating ensemble approaches. However, existing LLM ensembles focus on text generation or classification, leaving NER under-explored. In this context, this work proposes a novel three-step ensemble pipeline for zero-shot NER using similarly capable, locally run LLMs. Our method outperforms individual LLMs in four out of five Portuguese NER datasets by leveraging a heuristic to select optimal model combinations with minimal annotated data. Moreover, we show that ensembles obtained on different source datasets generally outperform individual LLMs in cross-dataset configurations, potentially eliminating the need for annotated data for the current task. Our work advances scalable, low-resource, and zero-shot NER by effectively combining multiple small LLMs without fine-tuning. Code is available at \url{https://github.com/Joao-Luz/local-llm-ner-ensemble}.

\end{abstract}
\section{Introduction}

Large Language Models (LLMs) have shown great capabilities in performing various Natural Language Processing (NLP) tasks requiring little to no labeled data due to their In Context Learning (ICL) capabilities \cite{zhao_survey_2025}. Notably, however, they still under-perform for token classification tasks such as Named Entity Recognition (NER) compared to smaller, specialized models \cite{wang_gptner_2023}. This performance gap is more apparent when using smaller LLMs designed to run on consumer GPUs instead of larger, often cloud-based, alternatives like OpenAI's ChatGPT, Anthropic's Claude, or Google's Gemini. Furthermore, the gap widens when performing such tasks for lower-resource languages such as Portuguese, in which these models notably under-perform when compared to resource-rich languages \cite{Garcia2024}.

Recently, new large language models are developed at a fast pace, with many of them being open-weights and available for local instantiation, such as the Meta's LLaMA family of models \cite{grattafiori_llama_2024}, Google's Gemma \cite{team_gemma_2024} and Microsoft's Phi \cite{abdin2024phi3technicalreporthighly}. However, even though many models have been made publicly available, no single model outperforms others for all different tasks, with them often complementing each other even within similar tasks \cite{jiang_llmblender_2023}. To that extent, using LLMs in ensembles presents itself as a means to achieve model-independent systems that can tap into different capabilities achieved by the various architectures and pre-training processes that each model possesses. When paired with the models' capabilities of performing tasks with no labeled examples, zero-shot LLM ensembles also stand out as scalable alternatives for tasks in low resource domains \cite{farr_llm_2024}, such as the case for many real-world applications of NER, especially for the Portuguese language. 

Previous methods of LLM ensembles rarely aim to address named entity recognition, often focusing on text generation \cite{jia_review_2024,jiang_llmblender_2023,jiang2024mixtralexperts} and text classification \cite{farr_llm_2024}. While some works aim to perform NER with LLM ensembles, they either fall short of successfully implementing a pipeline that shows performance gains when using different models \cite{nunes_ensemble_2025} or build pipelines that depend on the presence of larger, fine-tuned, and incidentally, more robust models to tend to more difficult cases \cite{li_improving_2025}. To the best of our knowledge, no work aims to aggregate LLMs of similar size and capabilities to perform zero-shot NER.

\begin{figure}[t]
    \centering
    \includegraphics[width=0.9\linewidth]{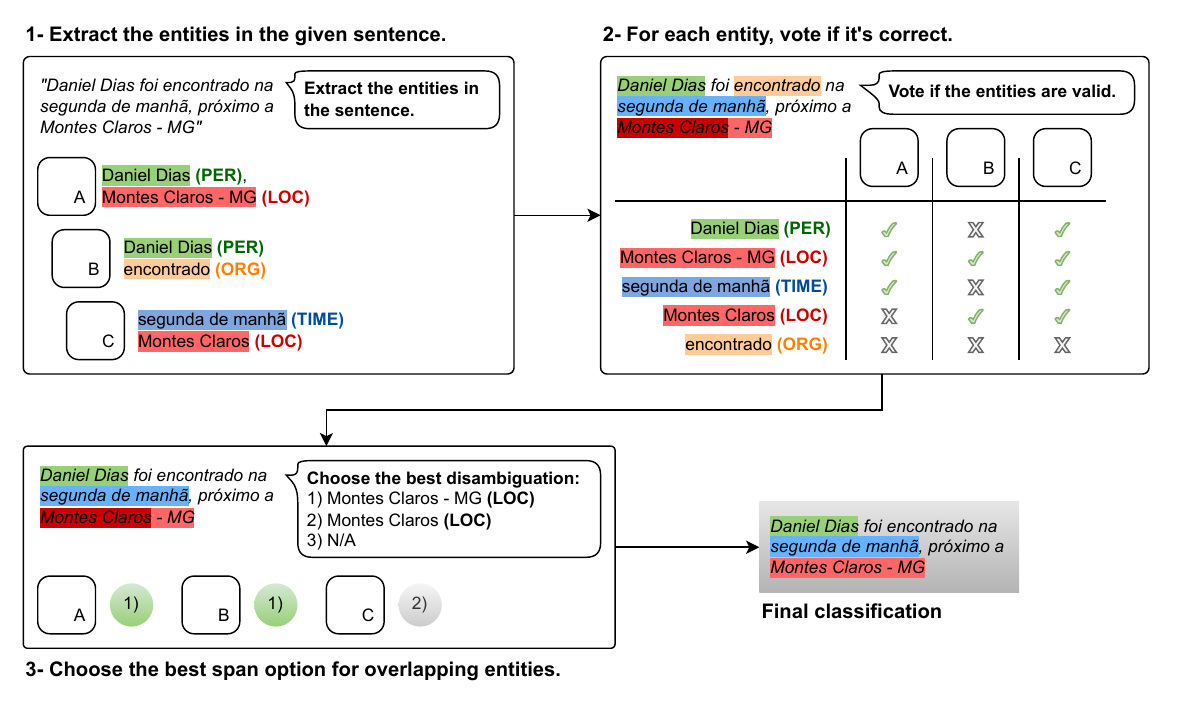}
    \caption{Pipeline for named entity recognition using the LLM ensemble. Three LLMs, A, B, and C, extract entities independently in step 1. In step 2, all LLMs vote if the entities mentioned are correct, and entities chosen by majority vote are, therefore, considered correct. Step 3 disambiguates the overlapping mentions (``\textit{Montes Carlos}'' and ``\textit{Montes Carlos - MG}'') by the LLMs through a multiple-choice prompt. The majority vote is then used to decide on the best fit for the entity in the spans.}
    \label{fig:pipeline}
\end{figure}

To address this gap, we outline a novel LLM ensemble pipeline focused on zero-shot named entity recognition intended to work with multiple, similarly sized, locally run models compatible with consumer GPUs. The pipeline, illustrated in Figure~\ref{fig:pipeline}, consists of three main steps: \textbf{extraction}, in which all models extract entities from a given document; \textbf{voting}, when all models decide, collectively, which entity mentions extracted in the previous step are correct; and \textbf{disambiguation}, a step to deal with possible overlapping entities deemed correct by the majority vote in the previous step. As the individual LLMs might perform differently for each pipeline task, different sets of models are considered for the different stages of the pipeline. In order to select the best configurations of models, we employ a heuristic that utilizes a small set of annotated data to evaluate the different combinations of LLMs and chooses the best-performing one.

We evaluated the ensemble method and compared its performance against individual models for five distinct Portuguese NER datasets, using zero-shot prompts for all pipeline steps. We found that the ensembles chosen by the proposed method performed better than any single LLM for four out of the five datasets. Moreover, we found that ensemble configurations obtained with different source datasets tended to perform as well or better than the ones obtained in the target dataset, which makes the pipeline feasible even in scenarios where no annotated data is available. In summary, this work's main contributions are:

\begin{itemize}
    \item Propose a novel pipeline for using multiple LLMs as an ensemble of zero-shot models for NER;
    \item Evaluate our method on 5 Portuguese NER datasets spanning different domains and entity granularities;
    \item Show that using ensembles obtained using annotated data from either the target task or a distinct source task tends to produce ensembles that perform zero-shot NER better than any single LLM.
\end{itemize}

\section{Related Work}

Ensemble methods with LLMs have recently become a popular research topic. These methods take advantage of the different models' diverse strengths while avoiding some of their specific weaknesses \cite{polikar_ensemble_2006}. A recent survey \cite{jia_review_2024} shows some of the most prominent ensembling strategies for LLMs used in the literature, which come in different varieties. Mix of Experts (MoE) models, such as Mixtral \cite{jiang2024mixtralexperts}, are composed of an ensemble of trainable weights selected during inference to produce the final output text. Fusion-based ensembling methods, such as LLM-Blender \cite{jiang_llmblender_2023}, introduce ways to combine the output of several pre-trained LLMs to produce a final text output, typically through reranking or voting, without depending on any training of the underlying LLMs. These approaches are primarily designed for generative tasks and may not trivially extend to structured prediction problems like NER, where span consistency and token-level granularity are important.

Routing \cite{chen_frugalgpt_2023,farr_llm_2024}, on the other hand, implements a mechanism to choose appropriate models for a given task input dynamically and, while applicable to text generation tasks, is simpler to be used for classification tasks. \cite{farr_llm_2024} propose a cascading chain of LLMs, where simpler inputs are processed by cheaper models and complex cases routed to more powerful ones. Simpler inputs are processed by the initial, simpler, and cheaper LLMs, while complex examples are forwarded to the more resource-intensive models. While cost-effective, this method assumes a predefined hierarchy of model capability and does not address scenarios where no single model in the chain performs optimally.

For NER specifically, \cite{li_improving_2025} uses fine-tuned models, as well as fine-tuned variations of API-based LLMs such as GPT-3.5, to perform entity recognition on medical domain data. They found that using LLMs together with other models in a majority vote format usually outperforms the individual base models. It is important to note that this work relies on the use of large, cloud-based models and large amounts of data to fine-tune the ensemble NER models.

For Portuguese NER, \cite{nunes_ensemble_2025} attempts to implement an ensembling method using the outputs of BERT \cite{devlin_bert_2019} sequence classifiers as inputs for a Mistral \cite{jiang2023mistral7b} large language model to perform aggregation of different entities. They query the model to decide on a final classification for a given span of the source sentence, providing information such as entity definitions and vote counts (number of BERT models that predicted such span as an entity.) The authors found that using this approach leads to performance losses compared to using the base classifiers alone. It is relevant to note, however, that their approach relies on a single model to perform the aggregation of information and does not employ multiple LLMs for the task. Furthermore, their analysis is limited to the legal domain for Portuguese NER.

\section{Proposed Method}

The proposed LLM ensemble pipeline\footnote{Available at \url{https://github.com/Joao-Luz/local-llm-ner-ensemble}} for entity recognition consists of three main steps: \textbf{entity extraction}, when each LLM of the ensemble is tasked with extracting entities from the given sentence; \textbf{voting}, when the LLMs must vote if the entities extracted by the entire ensemble in the previous step are valid or not for the sentence; and \textbf{disambiguation}, when LLMs are tasked with choosing the best disambiguation of overlapping entities obtained in the previous step. This section details each step of the pipeline illustrated in Figure \ref{fig:pipeline} .

\subsection{Entity Extraction}

Our ensemble approach begins with independent entity extraction by each LLM using a zero-shot prompt strategy adapted from \cite{xie_empirical_2023}. The prompt consists of the task definition entity type descriptions and queries the model to generate the list of entities in a sentence in JSON format. The LLMs' responses are then parsed, and each entity is validated in order to remove those with inconsistent types or those that can not be matched in the original sentence. Finally, all extracted entities from all of the ensemble's models are compiled into a unified record.

\subsection{Voting}

After extracting entities using the different LLM, each one is assigned to vote on whether the extracted entities are valid or not. It mimics a human annotation system, where multiple annotators might decide on the validity of entities annotated by different annotators in a team. Each extracted entity is used to construct a prompt that queries the LLM if the given entity mention is correct for its respective sentence. After each LLM takes a vote, we use the majority strategy to decide whether the entity is valid. This process generates a list of entities sourced individually from all models in the ensemble and validated by all of them.

\subsection{Disambiguation}

\begin{figure}[htpb]
    \centering
    \includegraphics[width=0.7\linewidth]{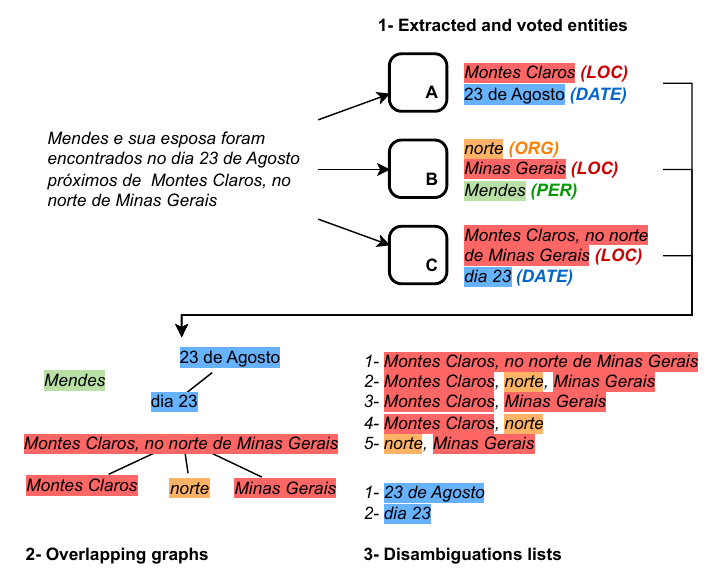}
    \caption{Process for constructing the disambiguation prompt.}
    \label{fig:disambiguation-prompt}
\end{figure}

When different models identify overlapping entities, we group these entities for disambiguation. This situation is illustrated in Figure \ref{fig:pipeline} for the entities ``\textit{Montes Claros}'' and ``\textit{Montes Claros - MG}''. A multiple-choice prompt is applied so the ensemble's models may decide which entities are more correct for the given sentence. The construction process for this prompt is illustrated in Figure \ref{fig:disambiguation-prompt} with a more in-depth example.

The disambiguation considers all mentions that passed the voting step, building an overlap graph where connected mentions form ambiguous groups. For example, non-overlapping mentions like ``\textit{Montes Carlos}'', ``\textit{norte}'', and ``\textit{Minas Gerais}'', shown in Figure \ref{fig:disambiguation-prompt}, still require disambiguation because they all share the common overlapping mention ``\textit{Montes Claros, no norte de Minas Gerais}''. The set of possible disambiguations for the span are used to build the multiple-choice prompt, which contains the five longest valid entity combinations from the overlapping span, plus an ``N/A'' option. Each LLM must select its preferred option, with the most voted choice becoming the final entity disambiguation and ties being resolved by selecting the option with the longest combined entity lengths. The final entity list for a given sentence combines the non-overlapping entities from voting with the resolved entities from the disambiguation step.

\subsection{Choosing Members of the Ensemble}

Our experiments revealed that having all LLMs participating in all pipeline steps might lead to worse performance than using some separately. Some models would, for example, under-perform in the extraction step, introducing much noise into the extracted entity list, which could be challenging to remove in the voting and disambiguation steps. Similarly, some LLMs might be either too permissive or too strict in the voting and disambiguation steps, which could negatively impact the final list of recognized entities. To address this, we employ a heuristic to choose optimal combinations of models to perform each of the steps in the pipeline.

First, we select a small sample of the labeled data to act as a development set. Then, we execute the pipeline for all possible combinations of available LLMs for each step. We use every possible combination for the extraction step. For the voting and disambiguation steps, we only use combinations of odd numbers of models, as we rely on majority voting. After obtaining the performance for every combination of LLMs per step, we select the configuration that obtained the best micro-F1 performance overall. 

\section{Experimental Setup}

\subsection{Datasets}

We selected five diverse and gold-standard datasets for NER in Portuguese in order to evaluate the ensemble pipeline. The datasets span from general to specific domains, with different levels of granularity for entity types. The selected datasets are listed in Table \ref{tab:datasets}.

\begin{table}[ht!]
\centering
\scriptsize
\caption{Datasets used for evaluation with their entity types and domains.}
\label{tab:datasets}
\begin{tabular}{lll}
\toprule
\textbf{Dataset} & \textbf{Domain} & \textbf{Entity Types} \\ \midrule
HAREM (Selective) \cite{santos_harem_nodate} & General & \begin{tabular}[c]{@{}l@{}}Person, Location, Time,\\ Organization, Value\end{tabular} \\ \midrule
LeNER-Br \cite{villavicencio_lener-br_2018} & Legal & \begin{tabular}[c]{@{}l@{}}Person, Organization, Location,\\ Time, Legislation, Jurisprudence\end{tabular} \\ \midrule
UlyssesNER-Br \cite{albuquerque_ulyssesnerbr_2022} & Legal & \begin{tabular}[c]{@{}l@{}}Date, Event, Government Org.,\\ Non-gov. Org., Political Party,\\ Other Product, Program Product, \\ System Product, Virtual Location,\\ Physical Location, Person-Group Position, \\ Individual Person, Person Position,\\ Law Foundation, Bill Foundation, \\ Nickname Foundation\end{tabular} \\ \midrule
GeoCorpus-2 \cite{consoli_embeddings_} & Geological & \begin{tabular}[c]{@{}l@{}}Eon, Era, Period, Epoch, Age, \\ Siliciclastic Sedimentary Rock, \\ Carbonate Sedimentary Rock, \\ Chemical Sedimentary Rock, \\ Organic Sedimentary Rock, \\ Brazilian Sedimentary Basin, \\ Basin Geological Context, \\ Lithostratigraphic Unit, Misc.\end{tabular} \\ \midrule
MariNER \cite{sarcinelli2025marinerdatasethistoricalbrazilian} & Historical & \begin{tabular}[c]{@{}l@{}}Person, Location, Date, Organization \end{tabular} \\ \bottomrule
\end{tabular}
\end{table}

For every dataset, we use the available train split to perform ensemble configuration selection and test split for testing. 

\subsection{Models}

As we aim to develop an ensembling method that works with LLMs with similar capabilities, the models must be of comparable size in the number of parameters. They also must be sufficiently diverse in order to be able to capture complementary information. To that extent, we decided to employ five different models, developed by different teams, with varied pre-training backgrounds and architectures. They are shown in Table \ref{tab:models}:

\begin{table}[!h]
    \scriptsize
    \centering
    \caption{Selected LLMs and their sizes.}
    \label{tab:models}
    \begin{tabular}{lrr}
        \toprule
        \textbf{Model} & \textbf{Parameters} & \textbf{Organization} \\
        \midrule
        LLaMA3\tablefootnote{\url{https://huggingface.co/meta-llama/Meta-Llama-3.1-8B-Instruct}} \cite{grattafiori_llama_2024} & 8B & Meta \\
        Qwen2\tablefootnote{\url{https://huggingface.co/Qwen/Qwen2-7B-Instruct}} \cite{yang_qwen2_2024} & 7B & Alibaba \\
        Gemma2\tablefootnote{\url{https://huggingface.co/google/gemma-2-9b-it}} \cite{team_gemma_2024} & 9B & Google \\
        Phi3\tablefootnote{\url{https://huggingface.co/microsoft/Phi3-Medium-128K-Instruct}} \cite{abdin2024phi3technicalreporthighly} & 14B & Microsoft \\
        Mistral\tablefootnote{\url{https://huggingface.co/mistralai/Mistral-7B-Instruct-v0.2}} \cite{jiang2023mistral7b} & 7B & Mistral AI \\
        \bottomrule
    \end{tabular}
    \label{tab:llm_comparison}
\end{table}
\subsection{Configurations}

In order to get more diverse outputs from the individual models, we first perform an initial calibration step to determine an optimal temperature value for each LLM. This step consists of the first two steps of the pipeline, done individually for each model and evaluated on the labeled sample of size 100 obtained from the available train set. We perform extraction at various temperature levels $t$, and voting using temperature $t=0$. The final overall micro-F1 scores are used to determine the ideal temperature for each model in the extraction stage. For simplicity, we run this calibration step using only the HAREM dataset and use the optimal temperature levels for the remaining sets: $t=0$ for LLaMA3, Mistral, and Phi3, $t=1.0$ for Gemma2 and $t=1.5$ for Qwen2. For voting and disambiguation, temperatures are kept at $t=0$.

To choose the best combinations of models using the previously mentioned heuristic, a sample of 100 random sentences is taken from each dataset's train split to act as the validation set, simulating a low-resource scenario. The same sampling is used in the calibration step for the HAREM dataset. Testing is performed in the entire test set available for each dataset.

\section{Results and Discussion}

\subsection{Main Results}

Table \ref{tab:main} shows the micro-F1 performance of the best ensemble configuration obtained via the heuristic on both the validation and test sets. The table also shows the performances of individual models, which were obtained from processing the entire pipeline using the same model for extraction, voting, and disambiguation. It also shows the performance of a fully supervised RoBERTa model \cite{liu2019robertarobustlyoptimizedbert}, trained on the full train splits of the datasets. To account for the models' stochastic natures, the reported results are the average of three runs of each experiment.

\begin{table}[htpb]
\scriptsize
\centering
\caption{Performance of best ensemble configurations and individual LLMs for each dataset. The gray-colored rows show the performance of individual models. \textbf{Bold} indicates best performance among LLM-based approaches.}
\label{tab:main}
\begin{tabular}{l|ccc|cc}
\toprule
\textbf{Dataset} & \textbf{Extraction} & \textbf{Voting} & \textbf{Disambiguation} & \textbf{Validation} & \textbf{Test} \\ \midrule
 & LLaMA3 & \multicolumn{1}{c}{Qwen2} & \begin{tabular}[c]{@{}c@{}}Phi3, Gemma2,\\ LLaMA3\end{tabular} & \textbf{0.617} & 0.591 \\
 & \multicolumn{3}{c|}{\cellcolor[HTML]{EFEFEF}Phi3} & \cellcolor[HTML]{EFEFEF}0.588 & \cellcolor[HTML]{EFEFEF}0.605 \\
 & \multicolumn{3}{c|}{\cellcolor[HTML]{EFEFEF}Gemma2} & \cellcolor[HTML]{EFEFEF}0.584 & \cellcolor[HTML]{EFEFEF}\textbf{0.609} \\
 & \multicolumn{3}{c|}{\cellcolor[HTML]{EFEFEF}LLaMA3} & \cellcolor[HTML]{EFEFEF}0.536 & \cellcolor[HTML]{EFEFEF}0.545 \\
 & \multicolumn{3}{c|}{\cellcolor[HTML]{EFEFEF}Qwen2} & \cellcolor[HTML]{EFEFEF}0.513 & \cellcolor[HTML]{EFEFEF}0.556 \\
 & \multicolumn{3}{c|}{\cellcolor[HTML]{EFEFEF}Mistral} & \cellcolor[HTML]{EFEFEF}0.378 & \cellcolor[HTML]{EFEFEF}0.439 \\
\multirow{-7}{*}{HAREM} & \multicolumn{3}{c|}{\cellcolor[HTML]{EFEFEF}RoBERTa} & \cellcolor[HTML]{EFEFEF}- & \cellcolor[HTML]{EFEFEF}0.874 \\ \midrule
 & Gemma2 & \begin{tabular}[c]{@{}c@{}}Phi3, Gemma2,\\ LLaMA3\end{tabular} & \begin{tabular}[c]{@{}c@{}}Phi3, Gemma2,\\ Mistral\end{tabular} & \textbf{0.580} & \textbf{0.541} \\
 & \multicolumn{3}{c|}{\cellcolor[HTML]{EFEFEF}Phi3} & \cellcolor[HTML]{EFEFEF}0.490 & \cellcolor[HTML]{EFEFEF}0.454 \\
 & \multicolumn{3}{c|}{\cellcolor[HTML]{EFEFEF}Gemma2} & \cellcolor[HTML]{EFEFEF}0.556 & \cellcolor[HTML]{EFEFEF}0.528 \\
 & \multicolumn{3}{c|}{\cellcolor[HTML]{EFEFEF}LLaMA3} & \cellcolor[HTML]{EFEFEF}0.510 & \cellcolor[HTML]{EFEFEF}0.461 \\
 & \multicolumn{3}{c|}{\cellcolor[HTML]{EFEFEF}Qwen2} & \cellcolor[HTML]{EFEFEF}0.369 & \cellcolor[HTML]{EFEFEF}0.370 \\
 & \multicolumn{3}{c|}{\cellcolor[HTML]{EFEFEF}Mistral} & \cellcolor[HTML]{EFEFEF}0.299 & \cellcolor[HTML]{EFEFEF}0.325 \\
\multirow{-7}{*}{LeNER-BR} & \multicolumn{3}{c|}{\cellcolor[HTML]{EFEFEF}RoBERTa} & \cellcolor[HTML]{EFEFEF}- & \cellcolor[HTML]{EFEFEF}0.914 \\ \midrule
 & \begin{tabular}[c]{@{}c@{}}Gemma2, LLaMA3,\\ Mistral\end{tabular} & Phi3 & \begin{tabular}[c]{@{}c@{}}Gemma2, LLaMA3,\\ Qwen2\end{tabular} & \textbf{0.523} & \textbf{0.313} \\
 & \multicolumn{3}{c|}{\cellcolor[HTML]{EFEFEF}Phi3} & \cellcolor[HTML]{EFEFEF}0.388 & \cellcolor[HTML]{EFEFEF}0.243 \\
 & \multicolumn{3}{c|}{\cellcolor[HTML]{EFEFEF}Gemma2} & \cellcolor[HTML]{EFEFEF}0.341 & \cellcolor[HTML]{EFEFEF}0.262 \\
 & \multicolumn{3}{c|}{\cellcolor[HTML]{EFEFEF}LLaMA3} & \cellcolor[HTML]{EFEFEF}0.274 & \cellcolor[HTML]{EFEFEF}0.187 \\
 & \multicolumn{3}{c|}{\cellcolor[HTML]{EFEFEF}Qwen2} & \cellcolor[HTML]{EFEFEF}0.221 & \cellcolor[HTML]{EFEFEF}0.187 \\
 & \multicolumn{3}{c|}{\cellcolor[HTML]{EFEFEF}Mistral} & \cellcolor[HTML]{EFEFEF}0.270 & \cellcolor[HTML]{EFEFEF}0.178 \\
\multirow{-7}{*}{GeoCorpus2} & \multicolumn{3}{c|}{\cellcolor[HTML]{EFEFEF}RoBERTa} & \cellcolor[HTML]{EFEFEF}- & \cellcolor[HTML]{EFEFEF}0.862 \\ \midrule
 & LLaMA3 & \begin{tabular}[c]{@{}c@{}}Phi3, Gemma2,\\ Qwen2\end{tabular} & \begin{tabular}[c]{@{}c@{}}Mistral, LLaMA3,\\ Qwen2\end{tabular} & \textbf{0.420} & \textbf{0.353} \\
 & \multicolumn{3}{c|}{\cellcolor[HTML]{EFEFEF}Phi3} & \cellcolor[HTML]{EFEFEF}0.271 & \cellcolor[HTML]{EFEFEF}0.324 \\
 & \multicolumn{3}{c|}{\cellcolor[HTML]{EFEFEF}Gemma2} & \cellcolor[HTML]{EFEFEF}0.281 & \cellcolor[HTML]{EFEFEF}0.334 \\
 & \multicolumn{3}{c|}{\cellcolor[HTML]{EFEFEF}LLaMA3} & \cellcolor[HTML]{EFEFEF}0.340 & \cellcolor[HTML]{EFEFEF}0.309 \\
 & \multicolumn{3}{c|}{\cellcolor[HTML]{EFEFEF}Qwen2} & \cellcolor[HTML]{EFEFEF}0.155 & \cellcolor[HTML]{EFEFEF}0.291 \\
 & \multicolumn{3}{c|}{\cellcolor[HTML]{EFEFEF}Mistral} & \cellcolor[HTML]{EFEFEF}0.215 & \cellcolor[HTML]{EFEFEF}0.279 \\
\multirow{-7}{*}{UlyssesNER} & \multicolumn{3}{c|}{\cellcolor[HTML]{EFEFEF}RoBERTa} & \cellcolor[HTML]{EFEFEF}- & \cellcolor[HTML]{EFEFEF}0.889 \\ \midrule
 & Gemma2 & \begin{tabular}[c]{@{}c@{}}Phi3, Gemma2,\\ LLaMA3\end{tabular} & \begin{tabular}[c]{@{}c@{}}Phi3, Mistral,\\ Qwen2\end{tabular} & \textbf{0.767} & \textbf{0.706} \\
 & \multicolumn{3}{c|}{\cellcolor[HTML]{EFEFEF}Phi3} & \cellcolor[HTML]{EFEFEF}0.667 & \cellcolor[HTML]{EFEFEF}0.664 \\
 & \multicolumn{3}{c|}{\cellcolor[HTML]{EFEFEF}Gemma2} & \cellcolor[HTML]{EFEFEF}0.734 & \cellcolor[HTML]{EFEFEF}0.699 \\
 & \multicolumn{3}{c|}{\cellcolor[HTML]{EFEFEF}LLaMA3} & \cellcolor[HTML]{EFEFEF}0.652 & \cellcolor[HTML]{EFEFEF}0.603 \\
 & \multicolumn{3}{c|}{\cellcolor[HTML]{EFEFEF}Qwen2} & \cellcolor[HTML]{EFEFEF}0.648 & \cellcolor[HTML]{EFEFEF}0.657 \\
 & \multicolumn{3}{c|}{\cellcolor[HTML]{EFEFEF}Mistral} & \cellcolor[HTML]{EFEFEF}0.547 & \cellcolor[HTML]{EFEFEF}0.551 \\
\multirow{-7}{*}{MariNER} & \multicolumn{3}{c|}{\cellcolor[HTML]{EFEFEF}RoBERTa} & \cellcolor[HTML]{EFEFEF}- & \cellcolor[HTML]{EFEFEF}0.922 \\
\bottomrule
\end{tabular}
\end{table}

For most datasets, using the ensemble method presents a performance gain compared to using a single model, although not achieving the same performances as the supervised models (albeit, while not requiring any labeled data). Interestingly, the ensemble gain is more pronounced for LeNER-BR, GeoCorpus2, and UlyssesNER datasets, where individual LLMs tended to perform the worst for the zero-shot NER task. A possible explanation for this is that these three datasets are annotated for domain-specific entities, with GeoCorpus2 having 13 distinct types, UlyssesNER having 16, and LeNER-BR having 6. Due to the high amounts of unique types and the amount of domain knowledge necessary to identify them, individual LLMs struggle to distinguish between them. However, they may complement each other's knowledge gaps to produce a final, more robust list of entities for the sentences.

HAREM, on the other hand, is the only dataset where the ensemble performed worse than individual models. This might be explained due to the fact that it is the only generic domain set, annotated with relatively simple entity types. For this reason, individual LLMs might perform well enough such that it is more difficult to identify an ensemble configuration that complements knowledge gaps like the previous sets. Though the ensemble performs better for MariNER, it is only by a small margin, with Gemma2 having a drop in performance of only 0.01 in micro-F1. Even though the dataset belongs to a historical domain, its entity types are still relatively simple compared to the more complex types of the other sets. It may explain the small performance gap between the ensemble and the individual LLMs.

Regarding the composition of the ensembles, only GeoCorpus2 shows a configuration with more than one LLM performing extraction, while all others have either LLaMA3 or Gemma2 as the only extractors. Looking at the models' individual performances, we see that LLaMA3 and Gemma2 are high-performing for most of the datasets and, given the importance of the extraction step (i.e., the subsequent steps may be interpreted as filtering of the extracted entities), it is to be expected that the best individual models are often used as extractors. Interestingly, Phi3 also performed well in most of the individual cases, being the second-best individual option for all sets except LeNER-BR. However, it did not show up as an extractor for any of the best ensemble configurations. Instead, Phi3 shows up for most of the voting and disambiguation ensembles, which indicates that it might not be a strong extractor but a reliable verifier of extracted entities.

\subsection{Ablation}

An ablation study is conducted to verify the effectiveness of the different steps of the pipeline. Table \ref{tab:ablation} shows the performance of the ensembles obtained in the main results when either the voting step is omitted, or the LLM disambiguation step is simplified by always choosing the largest option from the multiple-choice prompt.

\begin{table}[t]
\scriptsize
\centering
\caption{Performances of ensembles omitting different steps from the pipeline.}
\label{tab:ablation}
\begin{tabular}{lrr}
\toprule
\textbf{Dataset} & \textbf{No disambiguation} & \textbf{No voting} \\
\midrule
HAREM & 0.592 & 0.410 \\
LeNER-BR & 0.541 & 0.478 \\
GeoCorpus2 & 0.271 & 0.178 \\
UlyssesNER & 0.351 & 0.249 \\
MariNER & 0.705 & 0.621 \\
\bottomrule
\end{tabular}
\end{table}

The performance is similar for most datasets when using the simplified version of the disambiguation step. For HAREM, the performance is slightly better than that of using LLMs for disambiguation, while for LeNER-BR, UlyssesNER, and MariNER, the performance is only slightly lower. For GeoCourpus2, however, there is a considerable performance drop when not using the ensemble's output for this step. It indicates that using the largest option might be a suitable simplification of the pipeline, reducing the computational cost by reducing the overall amount of prompts processed. Nevertheless, GeoCorpus2's performance drop also shows that this extra prompting might be important in some settings.

The voting step, on the other hand, proved to be important to the ensemble's performance, as suppressing this step consistently lowers the performance of the final ensemble, with considerable drops in micro-F1 scores across all datasets. This may be explained by the disambiguation step becoming the only step for entity filtering, which can be difficult to perform in the multiple-choice format. Moreover, without the voting step, noisy entity mentions show up more frequently in the disambiguation prompts, making the disambiguation more difficult for the LLMs to process.

\subsection{Effectiveness of The Heuristic}

\begin{figure}[htpb]
    \centering
    \includegraphics[width=0.65\linewidth]{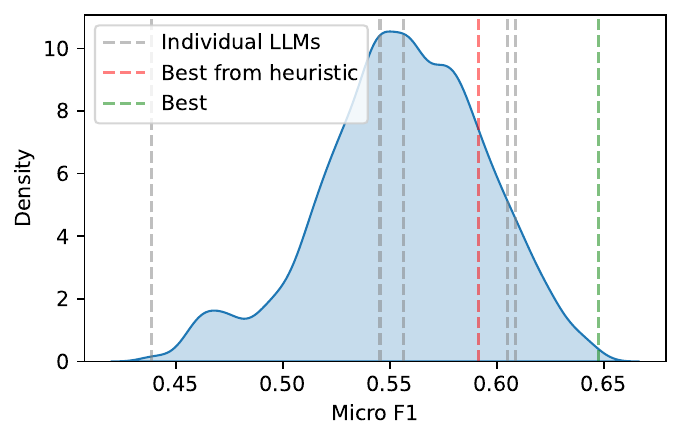}
    \caption{KDE plot of performances of all ensemble combinations for HAREM's test set. Performances for the individual LLMs, the best ensemble obtained from heuristics, and the best overall are highlighted.}
    \label{fig:histogram}
\end{figure}

Currently, an ensemble is chosen based on the performance of all combinations on a small annotated sample of the whole dataset, the validation set. However, Table \ref{tab:main} shows that the ensembles' performances on the validation set differ from the full test set. For that reason, we perform a study to verify the effectiveness of the chosen heuristic in selecting an ensemble configuration based on the small validation set. For this, all possible combinations of ensembles were tested for the entire test set of the HAREM dataset. HAREM was chosen here to analyze further the case where the ensemble underperformed against single LLM scenarios. Figure \ref{fig:histogram} compares the different configurations, showing the individual LLMs performances, the best configuration obtained through the heuristic, and the overall best ensemble.

Most ensemble configurations have a performance that falls close to 0.55, with two individual models (LLaMA3 and Qwen2) falling near that value. Interestingly, the best configuration possible, achieved by using Gemma2 for extraction, Phi3 for voting, and LLaMA3 disambiguation, achieves a performance greater than that of the ensemble obtained through the heuristic and the one obtained by Gemma2, the best individual LLM setting. It shows that the heuristic may not be the most effective way of obtaining the ensemble configuration or, at the very least, is highly dependent on the validation set size and its representativeness of the dataset as a whole.

\subsection{Cross Dataset ensembles}

\begin{figure}[htpb]
    \centering
    \includegraphics[width=0.7\linewidth]{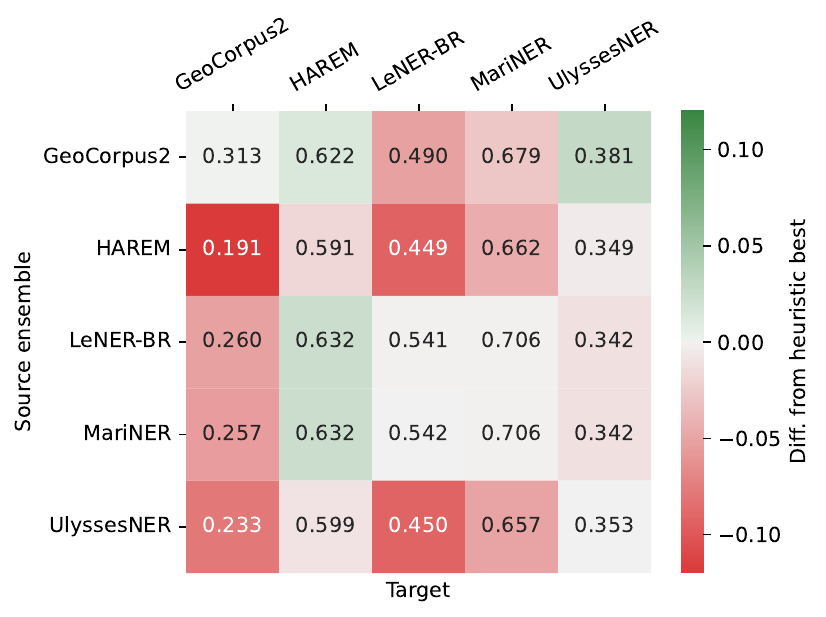}
    \caption{Cross dataset ensemble performances. Values are absolute micro-F1 performances, colours represent distance from the best configurations shown in Table \ref{tab:main}.}
    \label{fig:heatmap}
\end{figure}

We aim to assess how ensembles obtained from one dataset using the heuristic perform on another dataset by evaluating the ensemble pipeline in a scenario where no annotated data is available for the current task. The micro-F1 performances of ensembles obtained in a source dataset and tested in a different target set are shown in Figure \ref{fig:heatmap}.

The results show that using a different source dataset to obtain an ensemble configuration may perform better than any single LLM for most datasets. GeoCorpus2 is the only dataset in which this does not hold, with the second-best performing ensemble, obtained from LeNER-BR, still slightly under-performing Gemma2 in the test set. Notably, for both HAREM and UlyssesNER, it was possible to get an ensemble configuration that was better than the one obtained from the heuristic applied to the same dataset, which shows the capabilities of the ensemble pipeline of achieving high performances even in completely zero-shot settings, with no annotated data available for the current task.

\section{Conclusion and Future Work}

This study introduces a method to leverage multiple LLMs in ensembles to perform zero-shot named entity recognition for the Portuguese language. The pipeline is divided into three main steps: extraction, voting, and disambiguation. We evaluate this pipeline in five distinct NER Portuguese datasets, spanning multiple domains and annotation granularities. By evaluating the performance of all combinations of five unique LLMs across these steps for a small sample of labeled data, we can select a suitable ensemble configuration that performs better than any single LLM being used. We also show that ensemble configurations obtained in cross-dataset configurations may perform better than individual LLMs for four of the five datasets, eliminating the need for prior annotated data.

In future work, we aim to explore NER ensembling for languages other than Portuguese, as our pipeline is not dependent on this language. Other methods for aggregating the outputs of the LLM models may also be explored. One possibility would be to train a meta-classifier to aggregate the answers of the models similarly to what is usually done with ensemble stacking. Another alternative is to explore the use of larger models as judges in specific steps of the pipeline to use their processing capabilities to make final decisions.

\section{Limitations}

The use of LLM ensembles entails the instantiation of multiple GPU-intensive models for a single task. Naturally, this comes at the cost of high GPU memory requirements for parallel instantiation or long processing times to process the pipeline sequentially. Moreover, as we aim to obtain an optimal configuration of LLMs for each step of the pipeline, processing must be done for over 7,000 combinations to find the best-performing one for the small train set. To address this, we make heavy use of caching of previous answers for step 3 of the pipeline, but the processing time for all of these combinations still presents itself as a limiting factor for the proposed pipeline.

\section{Acknowledgements}

This study was financed in part by the Coordenação de Aperfeiçoamento de Pessoal de Nível Superior – Brasil (CAPES) – Finance Code 001.

%
%
%
\bibliographystyle{splncs04}
\bibliography{refs}
%




\end{document}